\documentclass[conference]{IEEEtran}
\IEEEoverridecommandlockouts

\usepackage[T1]{fontenc}

\usepackage{subcaption}
\usepackage[noadjust]{cite}
\usepackage{graphicx}
\usepackage{authblk}
\usepackage{hyperref}

\def\BibTeX{{\rm B\kern-.05em{\sc i\kern-.025em b}\kern-.08em
    T\kern-.1667em\lower.7ex\hbox{E}\kern-.125emX}}
    
\begin{document}

\title{Utilizing Novelty-based Evolution Strategies to~Train Transformers in~Reinforcement Learning}

\author[1,2]{Maty\'a\v{s} Lorenc}
\author[2]{Roman Neruda}
\affil[1]{Faculty of~Mathematics and~Physics, Charles University, Prague, Czech Republic}
\affil[2]{Institute of~Computer Science, Czech Academy of~Sciences, Prague, Czech Republic \authorcr lorenc@kam.mff.cuni.cz}

\maketitle

\begin{abstract}
In~this paper, we experiment with~novelty-based variants of~\mbox{OpenAI-ES}, the~\mbox{NS-ES} and~\mbox{NSR-ES} algorithms, and~evaluate their effectiveness in~training complex transformer-based architectures designed for~the~problem of~reinforcement learning, such~as~Decision Transformers. We also~test whether~we can accelerate the~novelty-based training of~these~larger models by~seeding the~training with~a~pretrained model. The~experimental results were mixed. \mbox{NS-ES} showed progress, but~it would clearly need many more iterations for~it to~yield interesting agents. \mbox{NSR-ES}, on~the~other hand, proved quite capable of~being straightforwardly used on~larger models, since its~performance appears as~similar between the~feed-forward model and~Decision Transformer, as~it is for~the~baseline objective-based \mbox{OpenAI-ES}.
\end{abstract}


\section{Introduction}

Reinforcement learning is considered possibly the~most difficult and~in~the~future hopefully the~most useful subfield of~machine learning, since~it mimics how~humans and~animals learn~\cite{RL}. Among~many approaches to~solving it~\cite{RL}, we can find those based on~computing a~gradient to~optimize the~objective, but~also others that are~derivative-free. Evolutionary algorithms~\cite{EA} are~one such class of~general derivative-free optimization algorithms that can be used to~solve this problem. Evolution strategies~\cite{ESIntro}, which belong to~this algorithmic family, have been proved to~be a~viable and~competitive alternative to~gradient approaches for~the~(deep) reinforcement learning~\cite{ESforRL}. Even though generally, the~gradient approaches have better sample utilization, the~evolution strategies -- just as~many of~their cousins from~the~family of~evolutionary algorithms -- are greatly parallelizable, which~helps them overcome this limitation. Another benefit of~using an~evolutionary algorithm for~reinforcement learning is that~they have better exploration of~possible solutions; therefore, agents trained using evolution strategies are~usually more diverse than~those trained by~gradient-based algorithms~\cite{DRLvsES,Combination}.

This strong exploration can be further enhanced by~incorporating techniques such as~novelty search~\cite{NS1,NS2}, where~we search for~previously unseen solutions. And~we can even combine the~novelty with~the~objective and~obtain quality-diversity algorithms~\cite{QD}. A~fairly simple, yet highly effective examples of~such algorithms for reinforcement learning are \mbox{NS-ES} and~\mbox{NSR-ES}~\cite{NS-ES}, both being variants of objective-based \mbox{OpenAI-ES}~\cite{OpenAI-ES}.

On~a~different note, a~transformer architecture \cite{Transformer} has recently become the~preferred solution in~the~field of~neural networks and~supervised learning for an ever-growing range of~problems, be~it its~original task of~language processing~\cite{Transformer} or~an~image processing task~\cite{VisualTransformer}. In~particular, there~have been efforts to~reinterpret a~reinforcement learning as~a~sequence modeling problem, utilizing the~strengths of~transformers to~develop novel approaches for~tackling such~challenges. This has led to~models like the~Decision Transformer~\cite{DecisionTransformer} and~Trajectory Transformer~\cite{TrajectoryTransformer}. Initially introduced as~a~model for~offline reinforcement learning based on~supervised sequence prediction, its authors also claim that~the~Decision Transformer performs effectively in~traditional online reinforcement learning tasks as~well.

In~our~previous work~\cite{Previous}, we subjected the~combination of~the~\mbox{OpenAI-ES} and~the~Decision Transformers to~experiments testing the~ability of~derivative-free algorithms to~train this more complicated and~larger transformer architecture, compared to~the~simple feed-forward models that had been experimented with before. The~evolution strategy proved to~be a~viable method to~train the~transformer in~this~setting. Still, the~\mbox{OpenAI-ES} employs a~clear fitness guidance for training, so~as~a~next step, we decided to~test whether the~novelty still provides enough training signals even for~these larger models, and~hence~we conducted experiments with \mbox{NS-ES} and~\mbox{NSR-ES} testing their capability to~train Decision Transformers.

The~contribution of~this paper thus~consists in~demonstrating the~ability of~the~aforementioned novelty-based approaches to~train agents whose~policies are based on~a~transformer architecture in~a~reinforcement learning setting.\footnote{Our~code, together with~the~data from~all the~conducted experiments can be found on~GitHub repository on~the~following~link: \url{https://github.com/Mafi412/Novelty-based-Evolution-Strategies-and-Decision-Transformers}}

In the~following section, we present the~background for our experiments. In~Section~\ref{Experiments}, we introduce our experiments and~show their results, whereas in~Section~\ref{Discussion}, we discuss those results. We then conclude the~paper in~the~last section.

\section{Background}

\subsection{Evolution Strategies}\label{ES}

\textit{Evolution strategies} are quite a~successful family of~black-box nature-inspired derivative-free optimization algorithms. They were introduced as a~tool for~solving high-dimensional continuous-valued problems~\cite{ESIntro}. The evolution strategies work with~a~population of~real-valued vectors (called \textit{individuals}). In~each iteration (\textit{generation}), they derive a~new set of~individuals by perturbating (\textit{mutating}) the~original population; the~new set is then evaluated with respect to~a~given objective function (\textit{fitness function}), and~a~new generation is formed based on~these new individuals taking into~account their objective function value (\textit{fitness} or \textit{fitness value}).

To~apply an~evolution strategy as~a~reinforcement learning algorithm, a~correspondence between the~individuals in~the~algorithm and~the~reinforcement learning agents represented by~a~neural network is drawn using the~network's weights. The~weights are vectorized and~the~resulting vectors of~real numbers are then used as~the~individuals for~the~algorithm. The~fitness of~each~individual is then~defined as~the~average cumulative reward -- called return -- of~the~corresponding agent over~multiple episodes. Various evolution strategy algorithms have been proposed for~this~purpose~\cite{ESforRL}.

This reinforcement learning approach comes with~certain drawbacks. For~instance, computing an~individual's fitness necessitates running entire episodes. Additionally, its sample efficiency is relatively low compared to~gradient-based methods; in~other words, gradients usually allow us to~extract more information from~a~timestep or~an~episode -- a~sample.

Nevertheless, evolution strategies also~have many~advantages. Numerous evolution strategy algorithms are~highly parallelizable, and~a~significant portion of~the~research in~the~field has been concentrated on~this~characteristic. As~a~result, we have algorithms that achieve a~linear performance improvement as~the~computing power increases~\cite{OpenAI-ES}. Furthermore, since evolution strategies are derivative-free, they allow optimization not~only of~conventional smooth neural networks but~also~of~models that~include discrete subfunctions or~other non-differentiable components. Another benefit is that~compared to~gradient-based algorithms, the~evolution strategies have superior exploration~\cite{DRLvsES,Combination}.

This~inherently good exploration can be further vastly increased by~utilizing \textit{novelty}~\cite{NS1,NS2}, which~basically means searching for~novel, previously unseen behavior, as~compared to~the~classical fitness-based approach that~seeks high-performing behaviors. The~novelty can either completely replace the~objective function, which~yields us so-called \textit{novelty search} algorithms~\cite{NS1,NS2} -- open-ended algorithms suitable when~the~rewards of~an~environment are not~informative enough, when~they are~deceptive, or~when~they are~hard to~reasonably specify -- or~it can be used to~complement the~rewards, yielding us \textit{quality-diversity} algorithms~\cite{QD}.

In~order to~compute the~novelty, we need to~first define what~we understand by~\textit{behavior} in~a~given environment. It should describe, what an~agent does in~the~environment. Then, we need to~specify a~distance metric between two behavior characteristics, that will tell us how~similar they are. We also~need to~store encountered behaviors in~a~\textit{behavior archive}. The~novelty of~an~individual is then~computed as~the~average distance of~its~behavior characteristic from~its~$k$ nearest neighbors in~the~archive.

In this paper, we will be working with~\mbox{\textit{NS-ES}} and~\mbox{\textit{NSR-ES}}~\cite{NS-ES}, two variations of~the~\mbox{\textit{OpenAI-ES}}~\cite{OpenAI-ES} algorithm utilizing novelty. The~first one is a~pure novelty search algorithm; the~second one belongs to~the~quality-diversity algorithms. It will be beneficial for~us to~first understand the~\mbox{OpenAI-ES} and~then extend this~algorithm into~the~two tested in~this~paper.

\mbox{OpenAI-ES} is a~representative of~Natural evolution strategies~\cite{NES}. It models the~population as~a~probability distribution over~the~agent’s (neural network’s) parameters, specifically a~Gaussian distribution. The~mean of~this~distribution serves as~the~candidate solution to~the~given problem, while~new offspring are generated by~sampling from~it each generation. These offspring are then evaluated, and~their performance is used to~update the~distribution’s parameters (in~our~case, only~the~mean) to~improve the~expected fitness of~future samples. This update follows an~approximation of~the~natural gradient (whence~the~name of~the~algorithmic family). In~our~case, the~natural gradient approximation is achieved by~renormalizing (rescaling) the~update based on~uncertainty. In~general, computing the~natural gradient would involve inverting a~so-called Fisher information matrix and~applying it to~the~gradient estimate. However, as~shown before in~the~literature~\cite{GradientVariants}, when~deriving parameter updates from~a~Gaussian distribution with~uniform variance across all~parameters -- just~as~we do -- dividing by~the~variance (i.e., rescaling with~respect to~uncertainty) yields a~similar effect and~achieves a~similar result. The~algorithm is also~designed for~highly efficient parallelization, minimizing interprocess communication. For~further details or~a~discussion of~the~design choices, we refer our~readers to~the~original paper~\cite{OpenAI-ES}.

\mbox{NS-ES} and~\mbox{NSR-ES} differ from~\mbox{OpenAI-ES} just in~a~few things. First, they maintain a~metapopulation of~several distributions (represented by their means) serving as~distinct populations. Only~the~behaviors of~the~distribution means are added to~the~archive. A~member of~the~metapopulation to~be improved in~a~given iteration is chosen proportionally to~its~current novelty. In~\mbox{NS-ES}, wherever~the fitness would be used, the~novelty (computed with~respect to~the~current state of~the~behavior archive) is used. As~for~the~\mbox{NSR-ES}, the~fitness is combined with~the~novelty by~averaging the~two.

\subsection{Transformers}\label{transformers}

\textit{Transformers} are currently the~state-of-the-art sequence-to-sequence neural architecture utilized for numerous tasks of~supervised learning~\cite{Transformer,VisualTransformer}. As~a~rule of~thumb, they appear to~possess strong generalization capabilities; the~greater the~larger the~model employed. However, achieving these impressive results also~demands a~substantial amount of~training data.

The~most important component, to~which the~transformers owe their~success, is a~self-attention layer, which~is used repeatedly throughout~the~network. For~each input sequence element, the~self-attention constructs a~"key", a~"query", and~a~"value". Next, an~$i$-th output element is obtained as a~linear combination of~all~values, with~each~value weighted in~proportion to~the~product of~the~query at~the~given ($i$-th) position and~the~key associated with~the~value. Thus, this integrates information from the~entire input sequence to~generate each~individual element of~the~output, as again expressed in~the~following equation.

\[ output_i = \sum_{j=1}^n \mbox{softmax}\left( query_i^T \cdot all\_keys \right)_j \cdot value_j \]

We can apply a~mask and~for~each~element conceal the~portion of~the~input sequence which~follows it. This ensures that~only~preceding information is used to~derive the~output element. This~technique is known as~a~causal masking, and~a~transformer which~implements it is referred to~as~a~causal transformer.

In~the~context of~reinforcement learning, the~goal is for~the~agent to~select actions at~each~timestep that~maximize its~return. This, however, can also~be viewed as~a~sequence modeling problem. Consequently, the~state-of-the-art architecture for~processing sequences, the~transformer, naturally comes into~play. This led to~the~introduction of~the~Decision Transformer~\cite{DecisionTransformer}.

The~core idea is that~the~agent's policy should produce an~action not~solely based on~the~most recent observation, but~rather on~the~entire history (or~the~portion that~fits within~the~context window) of~past observations and~actions. To~influence the~agent's performance, a~conditioning on~the~return-to-go was introduced, which~represents the~desired return from~a~particular timestep until~the~end of~the~episode.

\begin{figure}[t]\centering
\includegraphics[width=0.48\textwidth]{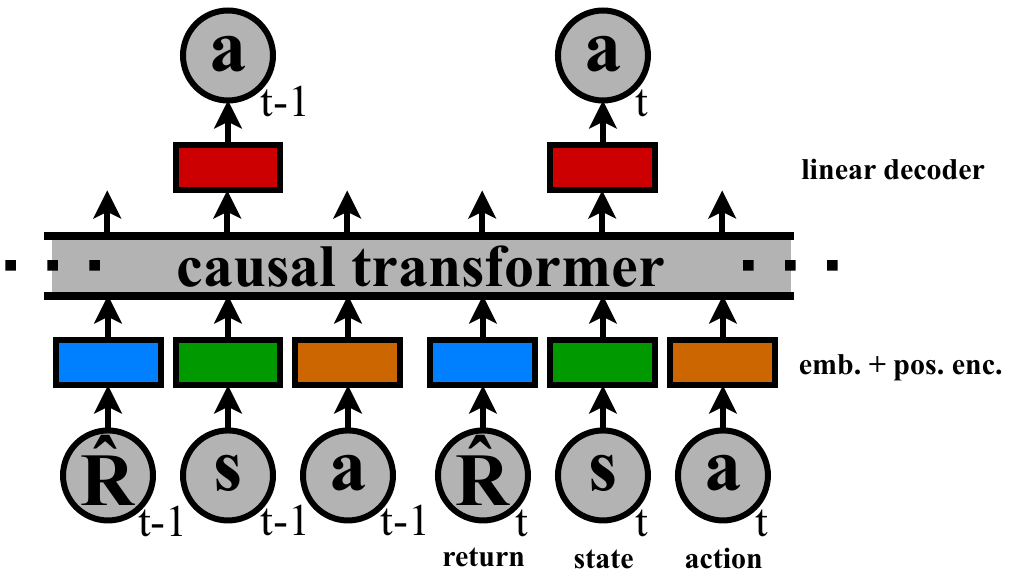}
\caption{Decision Transformer architecture \protect\cite{DecisionTransformer}}
\label{DT_architecture}
\end{figure}

Now, let us examine~the~proposed architecture itself. It consists of~a~causal transformer; embeddings for returns-to-go, observations (states of~the~environment), and~actions; position encoder; and~a~linear decoder to~transform the~output of~the~transformer into~actions, as shown in~Fig.~\ref{DT_architecture}.

At~each~timestep, the~model receives a~sequence consisting of~past triplets: return-to-go, observation, and~the~action taken. We then append the~current return-to-go and~observation, along~with~a~placeholder for~the~action that~has not been performed yet. Each~sequence component is passed through~its~respective embedding, the~positional encoding is added, and~the~sequence is then~processed by~the~transformer. The~element of~the~output sequence corresponding to~the~last input state is decoded to~determine the~next action to~perform.

A~key distinction from~a~typical transformer is that, for~each~timestep, all~parts of~the~triplet (return-to-go, observation, and~action) share the~same positional encoding. In~contrast, a~standard transformer assigns a~unique positional encoding to~each~element in~the~input sequence.

The~return-to-go values are generated in~a~recursive manner. For~the~first timestep, the~user provides the~initial return-to-go, which~represents the~desired performance (i.e., the~target return). For~all~subsequent timesteps, the~return-to-go is calculated by~subtracting the~reward received in~the~previous timestep from~the~return-to-go of~that~timestep.

\section{Experiments}\label{Experiments}

In~our~previous work~\cite{Previous}, we tested the~ability of~\mbox{OpenAI-ES} to~train agents with~their~policy constituted by~a~Decision Transformer. We also proposed a~method for~aiding this~evolution strategy in~training large models using first a~pretraining of~the~large model in~a~form of~a~behavior cloning towards some smaller, easily trained, yet~possibly weaker model. The~results were promising. The~algorithm was mostly capable of~training Decision Transformers, even~without~the~pretraining.

A~logical next step is to~test whether the~novelty described in~the~previous section provides us with enough information to~train these~larger models. Therefore, we extended our~implementation of~\mbox{OpenAI-ES} into~implementations of~\mbox{NS-ES} and~\mbox{NSR-ES} and~proceeded to~test these~novelty search and~quality-diversity algorithms in~the~MuJoCo~\cite{MuJoCo} Humanoid environment using OpenAI Gym~\cite{Gym}. For~details of~the~implementation, we refer our~reader to~the~original papers for~\mbox{OpenAI-ES}~\cite{OpenAI-ES} and~for~\mbox{NS-ES} / \mbox{NSR-ES}~\cite{NS-ES} regarding overall details, and~to~our~previous paper~\cite{Previous} regarding a~few tweaks in~the~implementation and~their~justifications, as~our~current code simply extends the~previous implementation into a~novelty-utilizing form.

Because training a~larger model -- which~the~Decision Transformer is -- is time-intensive, we chose to~test the~performance of~the~algorithms only~in~the~Humanoid environment. It should be a~suitable representative of~MuJoCo environments, as~it is the~most complex and~challenging among~the~standard ones.

As mentioned in Section \ref{ES}, utilizing novelty requires a behavior characteristic. We use the standard one for MuJoCo environments: the behavior characteristic of an agent is defined as its final position in the environment -- that is, the $x$ and $y$ coordinates at which the agent concludes the episode. The distance between two such characteristics is defined as a classical Euclidean distance between the two points.

For~both the~inspected algorithms, we first conducted a~replication experiment of~the~original paper \cite{NS-ES} and~a~correctness check of~our~implementation using a~classical feed-forward network, which then also served as~a~baseline for~follow-up experiments. We then proceeded with~experiments on~Decision Transformers and~concluded by~testing whether~the~pretraining helps to~accelerate the~novelty-based training of~these~larger models.

In~all the~experiments with Decision Transformers, we used the~same values for~the~model's hyperparameters that were used in~the~original paper~\cite{DecisionTransformer} for~Humanoid environment. Just to compare the~model sizes of the~feed-forward model and~the~Decision Transformer, which were used during our~experiments -- and~which correspond to~the~models used in~original papers~\cite{NS-ES,DecisionTransformer} -- the~feed-forward model has 166\,144 parameters and~the~Decision Transformer has 825\,098 parameters.

For~all the~experiments, we conducted ten runs of~the~training. For~each run of~each of~the~experiments, 300 workers were used utilizing 301 CPU cores (with one being a~master handling synchronization, evaluation, and~saving the~agent). The~desired returns passed to~all the~Decision Transformer models at~the~beginning of~each episode were 7000. However, it is important to~note that~in~the~original Decision Transformer paper~\cite{DecisionTransformer}, rewards and~returns-to-go are scaled down by~a~factor of~1000 in~MuJoCo environments. To~stay consistent with~this~approach, we have adopted the~same rescaling. All~figures showing fitness progression during~the~evolution training in~the~Humanoid environment are based on~these~scaled returns, as~this is the~format in~which the~values are logged during~training. To~retrieve the~actual fitness values as~produced by~the~original Humanoid environment, one~simply needs to~multiply the~reported numbers by~1000.

Unless stated otherwise in~the~individual experiment descriptions, all the~remaining hyperparameter values can be found as~default values of~the~arguments in~our codebase.

\begin{figure*}[p]

	\begin{subfigure}{0.47\textwidth}
        \includegraphics[width=\textwidth]{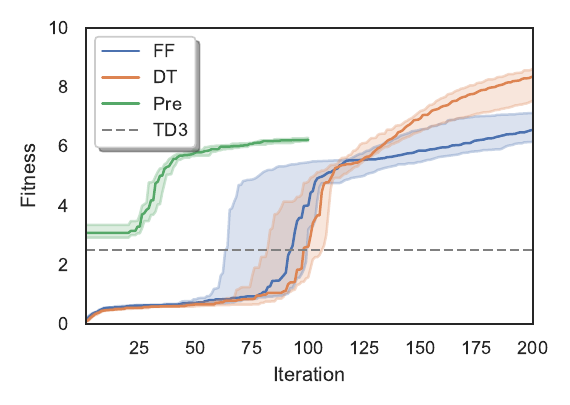}
        \caption{\mbox{OpenAI-ES} - Evaluation results}
        \label{ES_eval_figure}
    \end{subfigure}
    \hfill
	\begin{subfigure}{0.47\textwidth}
        \includegraphics[width=\textwidth]{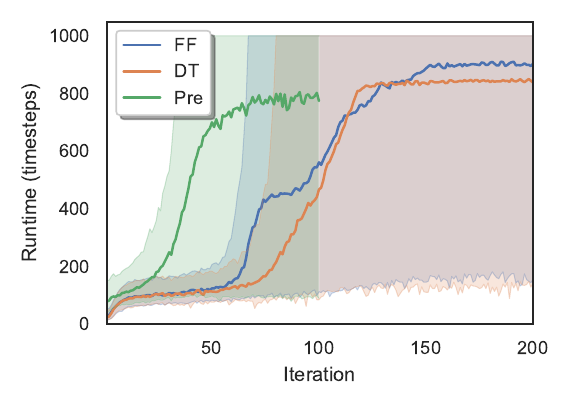}
        \caption{\mbox{OpenAI-ES} - Runtimes}
        \label{ES_runtime_figure}
    \end{subfigure}

	\begin{subfigure}{0.47\textwidth}
        \includegraphics[width=\textwidth]{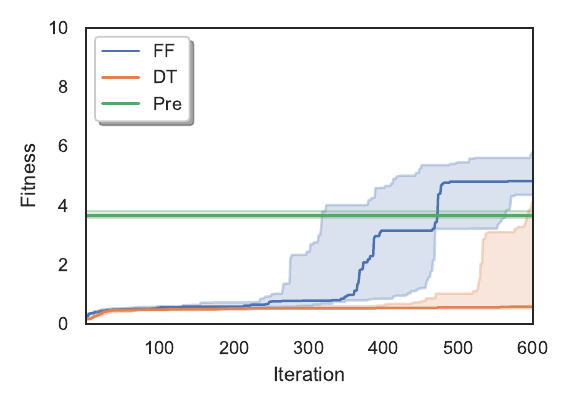}
        \caption{\mbox{NS-ES} - Evaluation results}
        \label{NS_eval_figure}
    \end{subfigure}
    \hfill
	\begin{subfigure}{0.47\textwidth}
        \includegraphics[width=\textwidth]{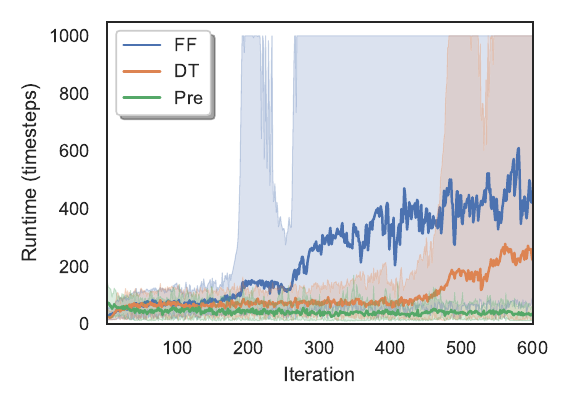}
        \caption{\mbox{NS-ES} - Runtimes}
        \label{NS_runtime_figure}
    \end{subfigure}

	\begin{subfigure}{0.47\textwidth}
        \includegraphics[width=\textwidth]{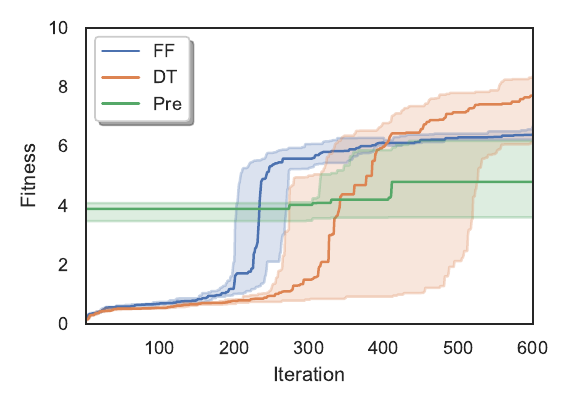}
        \caption{\mbox{NSR-ES} - Evaluation results}
        \label{QD_eval_figure}
    \end{subfigure}
    \hfill
	\begin{subfigure}{0.47\textwidth}
        \includegraphics[width=\textwidth]{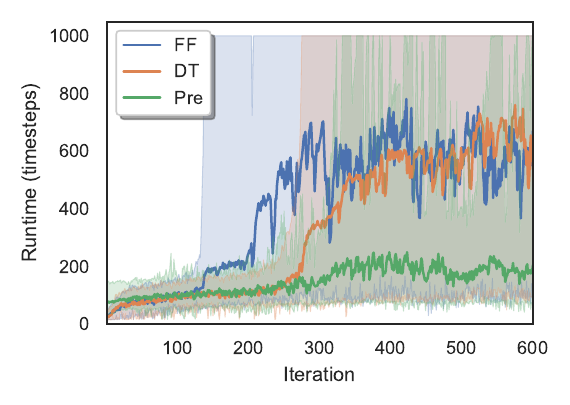}
        \caption{\mbox{NSR-ES} - Runtimes}
        \label{QD_runtime_figure}
    \end{subfigure}

\caption{\mbox{OpenAI-ES}, \mbox{NS-ES} and~\mbox{NSR-ES} used on~a~simple feed-forward model~(FF), a~Decision Transformer~(DT), and~a~pretrained Decision Transformer~(Pre) for~a~MuJoCo Humanoid simulation in~ten runs of~each of~the~experiments.
Fig.~\ref{ES_eval_figure}, \ref{NS_eval_figure}, and~\ref{QD_eval_figure} show solution evaluation fitness and~Fig.~\ref{ES_runtime_figure}, \ref{NS_runtime_figure}, and~\ref{QD_runtime_figure} show runtimes of~population evaluation episodes.
The~data from~the~ten runs are aggregated together: For the evaluation results, the median and quartiles are shown, whereas for the runtimes (where we have a lot more values per iteration), the~mean values and~percentile intervals (97,5\,\%) are displayed.
The~TD3 horizontal line in~Fig.~\ref{ES_eval_figure} shows the~best final average performance of~the~TD3 algorithm used on~a~Decision Transformer.
(In~Fig.~\ref{ES_eval_figure}, the~`Pre' line ends early because, for~\mbox{OpenAI-ES}, the~experiments with~pretrained agents were shorter than~those without~pretraining and~used only~100 iterations, compared to~200 for~the~others.)}
\label{Results_figure}
\end{figure*}

As~a~baseline for~the~novelty-based algorithms tested in~this~work, we present the~results that~the~purely objective-based algorithm, \mbox{OpenAI-ES}, was able to~achieve. These can be seen in~Fig.~\ref{ES_eval_figure} and~\ref{ES_runtime_figure}. Even~though there are no~purely gradient-based reinforcement learning algorithms utilizing novelty, it might still be interesting to~compare the~evolution-based approaches shown here with~some~gradient-based algorithm. Therefore, we show the~best final average performance achieved by~training a~Decision Transformer using TD3 algorithm during~roughly the~same wall-clock time as~the~\mbox{OpenAI-ES}. All~these objective-based results were adopted from~an~earlier paper~\cite{Previous}.

\subsection{NS-ES}

We start with~the~novelty search algorithm, \mbox{NS-ES}. We started with~a~replication experiment, where we used our~implementation of~this~algorithm to~train the~feed-forward model used in~the~original paper~\cite{NS-ES}. This also gave us a~good baseline for~our~experiments with~Decision Transformers. We then used the~same algorithm to~train the~Decision Transformer from~scratch; the~only difference was that because the~transformer is almost five times larger, we quadrupled the~size of~the~population the~algorithm works with. (This, of~course, increases the~number of~function evaluations -- episodes of~interactions with~the~environment -- by~a~factor of~four as~well.) The~last conducted experiment for~this~section was testing if~seeding the~training with~a~pretrained Decision Transformer accelerates the~process, as~the~training using just a~novelty signal is usually less~efficient than~following an~objective; hence~it might prove beneficial to~be able to~speed it~up, more~so when~training larger models. In~accordance with~previous research on~utilizing \mbox{OpenAI-ES} with transformers, no virtual batch normalization was used and~values of~learning rate and~noise deviation hyperparameters were both reduced to~$0.01$ when~utilizing the~pretraining. The~results of~these experiments can be seen in~Fig.~\ref{NS_eval_figure} and~\ref{NS_runtime_figure}.

Since our~objective shifts when~using novelty from~``teaching the~agent to~walk straight forward as~efficiently as~possible'' toward~simply ``teaching the~agent to~walk'', the~fitness is no~longer the~best representation of~the~agent's progress. (Still, it tells us how are we faring with~respect to~the~original objective.) Much~more informative are the~runtimes in~this~case, which~tell us how~long was the~agent able to~stay on~its~feet, and~hence~Fig.~\ref{NS_runtime_figure} is more important to~us now. There, we can observe that~the~progression in~training the~Decision Transformer when~not~using the~pretraining does indeed occur, but~much later than~when~training the~feed-forward model. That is in~stark contrast to~the~purely objective-based case shown in~Fig.~\ref{ES_eval_figure} and~\ref{ES_runtime_figure}, where the~difference between the~feed-forward and~Decision Transformer cases is not~so~significant. As~for~the~pretraining, it only~seems to~hurt the~training in~this~case, as~no~progress can be seen in~Fig.~\ref{NS_runtime_figure}. Still, in~Fig.~\ref{NS_eval_figure}, we can see that some progress is being made even with~respect to~the~fitness for~the~Decision Transformers, at~least without~the~pretraining.

In~Fig.~\ref{NS_runtime_figure}, we can see a~dent for~the~feed-forward data when, after reaching a~point where a~part of~the~population generated in~each~generation is able to~stay on~their~feet, this progress is reverted for~a~few iterations before~the~generated population regains this~ability. This is caused solely by~switching between various members of~the~metapopulation based on~their~current novelty throughout the~training, which is a~feature of~the~algorithm.

\begin{figure*}[t!]\centering
\includegraphics[width=\textwidth]{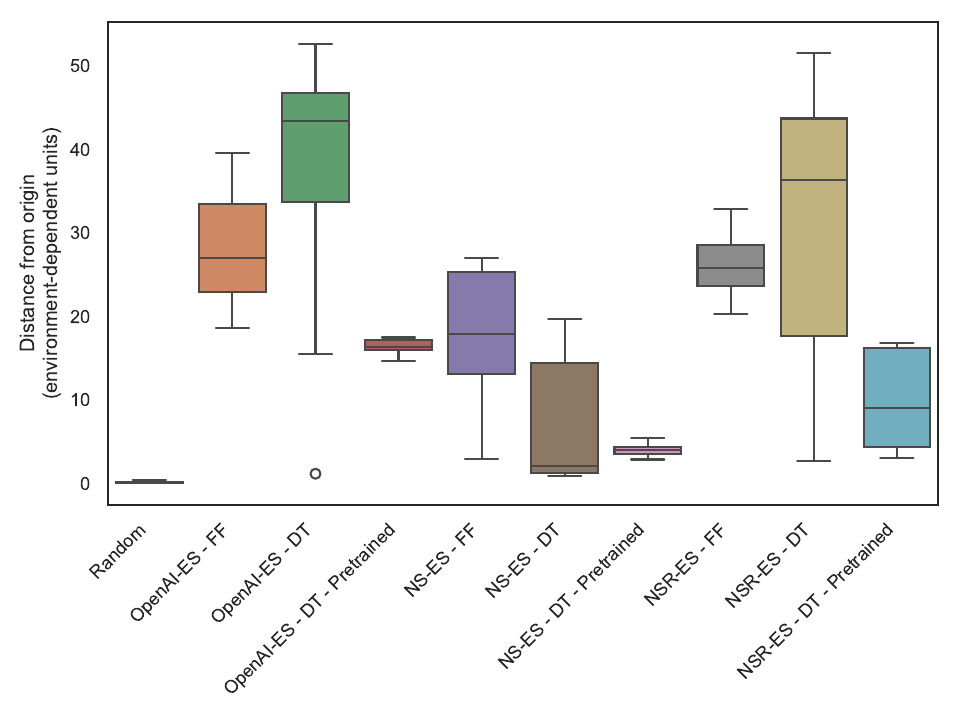}
\caption{Comparison of~the~average distances that~the~resulting agents trained by~the~examined algorithms were able to~travel from~their~starting positions in~the~Humanoid environment. Algorithms in~question are objective-based \mbox{OpenAI-ES}; novelty search \mbox{NS-ES}; quality-diversity \mbox{NSR-ES}; and~Random standing for~randomly initialized agents without~any~further training. When~an~algorithm works with~a~metapopulation, the~agent with~the~best average distance in~the~final metapopulation was chosen as~a~solution for~this~plot. Finally, for~each algorithm in~question (except~for~Random), results of~all three types of~experiments are plotted -- training of~a~feed-forward model~(FF) and~training of~a~Decision Transformer without~and~with~the~pretraining (DT and~\mbox{DT - Pretrained}, respectively). For~Random, only~the~feed-forward model was used.}
\label{Comparison_figure}
\end{figure*}

Of~course, one might ask about the~final performance of~our~trained agents with~respect to~our~above-declared objective: How~far are the~agents able to~walk after the~training? This can be seen in~Fig.~\ref{Comparison_figure}. We can clearly see that~even though the~novelty signal is capable of~somewhat training the~larger models, it would need much~more time to~achieve similar results as~with~the~smaller feed-forward models. And~we can see that the~pretrained Decision Transformers at~least fared better after~the~novelty search training than~the~random agents, but~this has to~be viewed in~context of~how~long the~training was and~that~the~pretrained agents were better before the~training.

\subsection{NSR-ES}

The~second set of~experiments was conducted with~the~quality-diversity \mbox{NSR-ES} algorithm. Again, a~replication experiment was performed -- our~implementation of~the~algorithm was used to~train the~feed-forward model. This gave us a~baseline for~our~further experiments with~the~transformers. And~in~the~same manner as~before in~the~previous subsection, we trained the~Decision Transformer from~scratch using \mbox{NSR-ES}, again~with~a~four-times larger population. We then concluded by~testing whether~seeding the~training with~a~pretrained transformer accelerates the~training. Again, just~as~in~the~previous case, when~training from~pretrained models, no virtual batch normalization was used and~values of~learning rate and~noise deviation hyperparameters were both reduced to~$0.01$. The~results of~these experiments can be seen in~Fig.~\ref{QD_eval_figure} and~\ref{QD_runtime_figure}.

This algorithm works considerably better on~the~Decision Transformers than the~novelty search \mbox{NS-ES} from~the~previous subsection; it is even sometimes capable of~further training the~pretrained models, even~though training the~pretrained models is still inferior to~training from~scratch, and~hence~offers no~benefits.

To~compare this~algorithm with~the~previous one with~respect to~the~final performance of~agents trained by~this~algorithm in~terms of~distance traveled, we can again take a~look at~Fig.~\ref{Comparison_figure}. We can see that~the~performance of~the~final agents based on~a~Decision Transformer this~algorithm yields is comparable to~the~performance of~such~agents trained by~the~objective-based \mbox{OpenAI-ES}. The~performance of~\mbox{NSR-ES} seeded with~a~pretrained Decision Transformer is similar to~\mbox{OpenAI-ES} -- when~the~training is successful -- but~it is not~so~reliable and~sometimes it fails to~train the~agent.

\section{Discussion}\label{Discussion}

In~the~previous section, we have seen that~\mbox{NS-ES} struggles to~train an~agent based on~a~Decision Transformer, and~that~it would need more compute to~succeed. This becomes even more obvious when~compared with~the~results yielded by~\mbox{OpenAI-ES}. Nevertheless, we should not forget that~novelty search is designed for~cases when~a~good reward function is not known, or~when the~rewards are deceptive and~lead the~agent to~a~local optimum, and~that it does not use any other signal than that of~novelty, which only tells it to~develop something new, but~does not tell it which "new" is preferred. And~in~such a~case, we can remark, it is possible to~train even larger models, however, they need a~good behavior characteristic -- which~is a~standard requirement for~novelty search algorithms -- and~an~ungodly amount of~computing power.

As~for~the~\mbox{NSR-ES}, it proved more successful. Although it does require more computation than~\mbox{OpenAI-ES}, it yields comparable results with~only a~threefold increase in~computation (thrice the iterations) while incorporating novelty and thus being theoretically capable of~overcoming local optima.

Finally, the~unfortunate pretraining. When~experimenting with~the~objective-based \mbox{OpenAI-ES}~\cite{Previous}, we noted that~using the~pretraining with~the~\mbox{OpenAI-ES} faced serious challenges and~proved to~be pretty much useless. Still, we hoped that~when~used in~the~context of~novelty-based training, it could insert a~previous knowledge and~accelerate the~training. However, this did not happen and~such attempts proved to~be futile.

A~possible solution, which~might be further explored in~a~future work, may be as~follows: We start by~training a~smaller, weaker model using \mbox{NS-ES} or~\mbox{NSR-ES}, saving the~behavior archive from~this~training. Then, we create a~new metapopulation of~larger models by~behavior cloning towards the~original metapopulation and~continue the~training with~this~new metapopulation and~the~saved behavior archive. This ensures that there is no reinventing the~wheel. In~other words, this makes sure that~the~simple behaviors already tried by~the~simple models are not developed again. We hypothesize that this was the~core reason for the nonfunctionality of~pretraining in~\mbox{NS-ES} and~\mbox{NSR-ES}, and~hence this could improve performance significantly. Another benefit is that~we could even keep and~further use the~virtual batch normalization data of~the~original metapopulation, as~the~inputs remain the~same and~we can use it already during the~behavior cloning. This eliminates~one of~the~problems of~the~utilization of~pretrained models identified during~experiments with~\mbox{OpenAI-ES}~\cite{Previous}. The~only remaining problem is that~the~new metapopulation might probably not be so~robust -- being trained by~a~gradient algorithm, by~a~behavior cloning -- and~hence reduced values for~hyperparameters influencing the~speed of~training would be required, which~would dampen the~progress of~further training. Perhaps this might be solved by~gradually increasing the~hyperparameters to~their~original values during the~training. If~this last bit could be resolved, we believe, this approach would outperform training the~large models from~scratch, and~so~it would help to~solve more complex problems requiring more complex models for~action selection.

\section{Conclusion}

We inspected the~ability of~novelty-based evolution strategies, in~our~case \mbox{NS-ES} and~\mbox{NSR-ES} -- a~novelty search and~a~quality-diversity algorithm, respectively -- to~train larger and~more complex models than~the~simple feed-forward ones that~are standard across the~reinforcement learning literature, like Decision Transformers. Although the~novelty search algorithm would require much~more computing power, the~quality-diversity algorithm proved to~be quite successful in~training these~bigger models.

We also suggested a~method for~utilizing previous knowledge -- previously trained simpler agents, respectively -- to~accelerate the~training of~those~larger models, yet~it proved to~be unsuccessful. Nonetheless, it allowed us to~formulate an~outline for~a~method that might possibly accelerate the~training, which, however, remains for~future work to~be finalized and~tested.

\bibliographystyle{unsrt}
\bibliography{bibliography}

\begin{thebibliography}{10}

\bibitem{RL}
Richard~S. Sutton and Andrew~G. Barto.
\newblock {\em Reinforcement Learning: An Introduction}.
\newblock The MIT Press, 2 edition, 2018.

\bibitem{EA}
Kenneth~A. De~Jong.
\newblock {\em Evolutionary Computation}.
\newblock The MIT Press, 2016.

\bibitem{ESIntro}
Ingo Rechenberg.
\newblock {\em Evolutionsstrategie — {O}ptimierung technischer {S}ysteme nach
  {P}rinzipien der biologischen {E}volution}.
\newblock Friedrich Frommann Verlag, Stuttgart-Bad Cannstatt, Germany, 1973.

\bibitem{ESforRL}
Paolo Pagliuca, Nicola Milano, and Stefano Nolfi.
\newblock Efficacy of modern neuro-evolutionary strategies for continuous
  control optimization.
\newblock {\em Frontiers in Robotics and AI}, 7, 2020.

\bibitem{DRLvsES}
Amjad~Yousef Majid, Serge Saaybi, Vincent Francois-Lavet, R.~Venkatesha Prasad,
  and Chris Verhoeven.
\newblock Deep reinforcement learning versus evolution strategies: A
  comparative survey.
\newblock {\em IEEE Transactions on Neural Networks and Learning Systems},
  35(9):11939--11957, 2024.

\bibitem{Combination}
Olivier Sigaud.
\newblock Combining evolution and deep reinforcement learning for policy
  search: A survey.
\newblock {\em ACM Trans. Evol. Learn. Optim.}, 3(3), September 2023.

\bibitem{NS1}
Joel Lehman and Kenneth~O. Stanley.
\newblock {\em Novelty Search and the Problem with Objectives}, pages 37--56.
\newblock Springer New York, New York, NY, 2011.

\bibitem{NS2}
Joel Lehman and Kenneth~O. Stanley.
\newblock Abandoning objectives: Evolution through the search for novelty
  alone.
\newblock {\em Evol. Comput.}, 19(2):189–223, jun 2011.

\bibitem{QD}
Justin~K. Pugh, Lisa~B. Soros, and Kenneth~O. Stanley.
\newblock Quality diversity: A new frontier for evolutionary computation.
\newblock {\em Frontiers in Robotics and AI}, 3, 2016.

\bibitem{NS-ES}
Edoardo Conti, Vashisht Madhavan, Felipe~Petroski Such, Joel Lehman, Kenneth~O.
  Stanley, and Jeff Clune.
\newblock Improving exploration in evolution strategies for deep reinforcement
  learning via a population of novelty-seeking agents.
\newblock In {\em Proceedings of the 32nd International Conference on Neural
  Information Processing Systems}, NIPS'18, page 5032–5043, Red Hook, NY,
  USA, 2018. Curran Associates Inc.

\bibitem{OpenAI-ES}
Tim Salimans, Jonathan Ho, Xi~Chen, and Ilya Sutskever.
\newblock Evolution strategies as a scalable alternative to reinforcement
  learning.
\newblock {\em arXiv}, 2017.

\bibitem{Transformer}
Ashish Vaswani, Noam Shazeer, Niki Parmar, Jakob Uszkoreit, Llion Jones,
  Aidan~N Gomez, \L{}ukasz Kaiser, and Illia Polosukhin.
\newblock Attention is all you need.
\newblock In {\em Advances in Neural Information Processing Systems},
  volume~30. Curran Associates, Inc., 2017.

\bibitem{VisualTransformer}
Alexey Dosovitskiy, Lucas Beyer, Alexander Kolesnikov, Dirk Weissenborn,
  Xiaohua Zhai, Thomas Unterthiner, Mostafa Dehghani, Matthias Minderer, Georg
  Heigold, Sylvain Gelly, Jakob Uszkoreit, and Neil Houlsby.
\newblock An image is worth 16x16 words: Transformers for image recognition at
  scale.
\newblock In {\em International Conference on Learning Representations}, 2021.

\bibitem{DecisionTransformer}
Lili Chen, Kevin Lu, Aravind Rajeswaran, Kimin Lee, Aditya Grover, Michael
  Laskin, Pieter Abbeel, Aravind Srinivas, and Igor Mordatch.
\newblock Decision transformer: Reinforcement learning via sequence modeling.
\newblock {\em arXiv preprint arXiv:2106.01345}, 2021.

\bibitem{TrajectoryTransformer}
Michael Janner, Qiyang Li, and Sergey Levine.
\newblock Offline reinforcement learning as one big sequence modeling problem.
\newblock In {\em Advances in Neural Information Processing Systems},
  volume~34, pages 1273--1286. Curran Associates, Inc., 2021.

\bibitem{Previous}
Matyáš Lorenc.
\newblock Utilizing evolution strategies to train transformers in reinforcement
  learning, 2025.
\newblock unpublished.

\bibitem{NES}
Daan Wierstra, Tom Schaul, Tobias Glasmachers, Yi~Sun, Jan Peters, and
  J\"{u}rgen Schmidhuber.
\newblock Natural evolution strategies.
\newblock {\em Journal of Machine Learning Research}, 15(27):949--980, 2014.

\bibitem{GradientVariants}
Thomas Pierrot, Nicolas Perrin-Gilbert, and Olivier Sigaud.
\newblock First-order and second-order variants of the gradient descent in a
  unified framework.
\newblock In Igor Farka{\v{s}}, Paolo Masulli, Sebastian Otte, and Stefan
  Wermter, editors, {\em Proceedings of the International Conference on
  Artificial Neural Networks (ICANN 2021)}, pages 197--208, Cham, 2021.
  Springer International Publishing.

\bibitem{MuJoCo}
Emanuel Todorov, Tom Erez, and Yuval Tassa.
\newblock {M}u{J}o{C}o: A physics engine for model-based control.
\newblock In {\em 2012 IEEE/RSJ International Conference on Intelligent Robots
  and Systems}, pages 5026--5033, 2012.

\bibitem{Gym}
Greg Brockman, Vicki Cheung, Ludwig Pettersson, Jonas Schneider, John Schulman,
  Jie Tang, and Wojciech Zaremba.
\newblock Open{AI} {G}ym.
\newblock {\em arXiv}, 06 2016.

\end{thebibliography}

\end{document}